\documentclass[10pt,twocolumn,letterpaper]{article}

\usepackage{cvpr}              

\usepackage{graphicx}
\usepackage{amsmath}
\usepackage{amssymb}
\usepackage{booktabs}
\usepackage{dcolumn,booktabs}
\usepackage{lipsum}
\usepackage[accsupp]{axessibility} 
\newcolumntype{d}[1]{D{.}{.}{#1}}

\usepackage[pagebackref,breaklinks,colorlinks]{hyperref}

\usepackage[capitalize]{cleveref}
\crefname{section}{Sec.}{Secs.}
\Crefname{section}{Section}{Sections}
\Crefname{table}{Table}{Tables}
\crefname{table}{Tab.}{Tabs.}

\def\cvprPaperID{3} 
\def\confName{CVPR}
\def\confYear{2024}

\hypersetup{colorlinks,linkcolor={blue},citecolor={blue},urlcolor={red}}

\begin{document}

\title{Does Head Pose Correction Improve Biometric Facial Recognition?}

\author{Justin Norman and Hany Farid\\
University of California, Berkeley\\
Berkeley CA, USA\\
{\tt\small justin.norman@berkeley.edu and hfarid@berkeley.edu}
}
\maketitle


\begin{abstract}
Biometric facial recognition models often demonstrate a significant reduction in accuracy when processing real-world images characterized by poor quality, non-frontal subject poses, and subject occlusions. We investigate whether targeted, AI-driven, head-pose correction and image restoration can improve recognition accuracy. Using a model-agnostic, large-scale, forensic evaluation pipeline, we assess the impact of three restoration approaches: 3D reconstruction (NextFace), 2D frontalization (CFR-GAN), and feature enhancement (CodeFormer). We find that naive application of these techniques degrades facial recognition accuracy. However, we also find that selective application of CFR-GAN combined with CodeFormer yields meaningful improvements.
\end{abstract}

\section{Introduction}
\label{sec:intro}

Fueled by the (often unverified) promise of accurate and automated identity verification, biometric facial recognition systems are increasingly deployed in both high-stakes commercial and forensic settings~\cite{phillips2018face,wang2021deep}. In controlled environments, modern face recognition models often report accuracies exceeding 95\%, which has led to public and institutional confidence in their deployment~\cite{smith2022ethical}. However, recent research has cast serious doubt on the generalizability of these results to real-world conditions~\cite{raji2020saving}, especially in forensic contexts. These systems are not as infallible, as often presumed, when tasked with matching images of faces captured under non-ideal conditions, such as occlusion, low quality, and low resolution.

A growing body of evidence suggests that images, often characterized by poor resolution, compression artifacts, occlusions, non-frontal and extreme poses, and inconsistent lighting, result in significant challenges to even state-of-the-art recognition systems. Under these conditions, recognition accuracy can drop precipitously, sometimes by more than $30$ percentage points compared to performance on high-quality benchmark datasets~\cite{norman2023forensicfacialrecognition}. These degradation factors, common in surveillance footage and law enforcement imagery, are not edge cases but emblematic of the forensic context.

Given this broader understanding of how and when facial recognition systems fail, a natural question emerges: Can targeted interventions yield meaningful improvements without overhauling the entire recognition pipeline? Rather than attempting to universally enhance all input data (a strategy shown to introduce artifacts and reduce accuracy in even ideal conditions)~\cite{deng2019arcface,kortylewski2019analyzing} or retrain the underlying recognition models, we explore whether selective, structure-aware preprocessing that preserves identity-bearing facial geometry rather than perceptual realism alone can offer a practical compromise.

One such treatment performs a high-fidelity 3D facial mesh reconstruction from a single image via the integration of a CNN encoder and ray tracing~\cite{dib2021practical,Dib_2021_ICCV}.  Separately, we evaluate a second full facial reconstruction process, which performs face frontalization and de-occlusion in a single model architecture, utilizing the CFR-GAN optimizer family and library~\cite{Ju_2022_WACV}. To further increase the fidelity of features within reconstructed and rectified images, we leverage the CodeFormer human subject restoration model~\cite{zhou2022towards}. We also introduce a new classifier to identify specific images that will benefit from these restoration and enhancement techniques.

Overall, we find that some targeted interventions are beneficial in specific circumstances. These benefits, however, are constrained by several factors, such as the ability to reliably detect forensic identification failures, overall image quality, facial orientation and the restoration technique used. As such, these interventions may only be applicable to specific use cases, have non-trivial requirements, and should not be considered a panacea.

\section{Related Work}
\label{sec:related-work}

The problem of face recognition under non-ideal conditions has a long history. Taigman et al.~\cite{taigman2014deepface} demonstrated near-human verification accuracy using deep convolutional features, while Schroff et al.~\cite{schroff2015facenet} introduced FaceNet with triplet-loss training for unified face embeddings. Deng et al.~\cite{deng2019arcface} advanced the field with ArcFace's additive angular margin loss, achieving state-of-the-art results on standard benchmarks. These models form the recognition backbone for our evaluation framework.

Pose variation remains a primary challenge for face recognition. Some progress has been made in increasing the availability of face data in non-standard poses. For example, Yaseen et al recently released a dataset containing multiple distinct gesture types~\cite{yaseen_2025_CVPR}. On the model side, Hassner et al.~\cite{hassner2015effective} proposed effective face frontalization using approximate 3D models. Tran et al.~\cite{tran2017drgan} introduced DR-GAN for disentangled pose-invariant representations, while Huang et al.~\cite{huang2017tpgan} developed TP-GAN for photorealistic frontal view synthesis combining global and local perception. These methods established frontalization as a viable strategy, though their impact on downstream biometric accuracy has received limited systematic evaluation.

Recent blind face restoration methods such as CodeFormer~\cite{zhou2022towards} leverage learned codebooks to recover high-quality facial detail from degraded inputs. However, as we previously demonstrated~\cite{Norman_2024_CVPR}, perceptual quality improvements do not reliably translate into biometric accuracy gains. Our work extends this investigation to head-pose correction specifically, using a controlled forensic evaluation framework rather than standard verification benchmarks.

\section{Forensic Facial Lineups}
\label{sec:forensic-lineups}

In previous work~\cite{norman2023forensicfacialrecognition}, we sought to assess the reliability of facial recognition technologies often used in real-world, high-stakes forensic scenarios, particularly when image quality is poor and/or faces are obscured. To do this, we developed a new, model-agnostic forensic evaluation framework for facial recognition systems. We then systematically tested the performance of widely used facial recognition models, ArcFace and FaceNet, using this process. The framework consists of the use of controlled image lineups to examine the impact of lineup composition and confounding factors like facial occlusions and variations in image quality.

These lineups are created by first extracting a vector embedding for each image in the dataset. The five most similar images to the source image -- based on embedding distance -- are added to the lineup. From there, a random probe image from the same identity as the source is added to the lineup. Using the vector embeddings extracted for each lineup image, the most similar image to the source is computed. If the most similar lineup image to the source is the probe image, then the evaluation for that lineup is considered to be successful; if not, then it represents a failure of the facial recognition system.

This prior evaluation demonstrated that recognition accuracy degrades substantially under realistic forensic conditions~\cite{norman2023forensicfacialrecognition}, motivating the restoration-based interventions explored here.

\section{Dataset}
\label{sec:datasets}

As in~\cite{norman2023forensicfacialrecognition}, we use a single, real-world dataset derived from the CASIA-WebFace dataset~\cite{yi2014learning}. This dataset consists of 491,414 images from 10,575 identities. Images vary in size, quality, pose, subject clothing, and environment. We performed some manual curation due to the initial quality of the dataset. This curation included the removal of duplicate images, incorrectly labeled images, and images that do not contain a human face. This dataset is significantly larger than the original dataset utilized in our previous research~\cite{norman2023forensicfacialrecognition,Norman_2024_CVPR}, which consisted of approximately 37,500 images from 7,500 identities.

\begin{figure*}[t!]
    \centering
    \small
    \setlength{\tabcolsep}{2pt}
    \setlength{\fboxsep}{0pt}  
    \setlength{\fboxrule}{2pt}  
    \renewcommand{\arraystretch}{1.0}

    \begin{tabular}{c@{\hspace{0.15cm}}|@{\hspace{0.15cm}}cccccc}
    \textbf{Source} & \multicolumn{6}{c}{Lineup}  \\
    \midrule
    \includegraphics[width=0.12\textwidth]{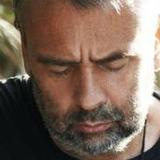} &
    \fcolorbox{green}{white}{\includegraphics[width=\dimexpr0.12\textwidth-2\fboxrule-2\fboxsep\relax]{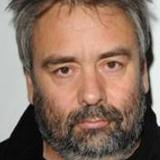}} &
    \includegraphics[width=0.12\textwidth]{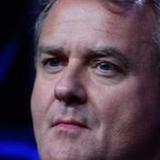} &
    \includegraphics[width=0.12\textwidth]{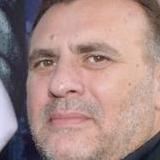} &
    \includegraphics[width=0.12\textwidth]{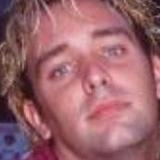} &
    \includegraphics[width=0.12\textwidth]{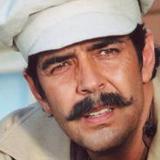} &
    \includegraphics[width=0.12\textwidth]{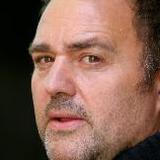} \\[-0.1cm]
    & \parbox[t]{0.12\textwidth}{\centering\small Clear} & & & & & \\[0.1cm]

    \includegraphics[width=0.12\textwidth]{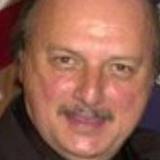} &
    \includegraphics[width=0.12\textwidth]{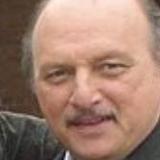} &
    \fcolorbox{red}{white}{\includegraphics[width=\dimexpr0.12\textwidth-2\fboxrule-2\fboxsep\relax]{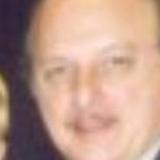}} &
    \includegraphics[width=0.12\textwidth]{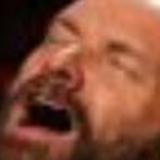} &
    \includegraphics[width=0.12\textwidth]{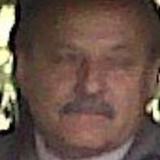} &
    \includegraphics[width=0.12\textwidth]{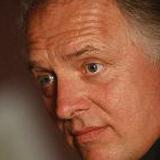} &
    \includegraphics[width=0.12\textwidth]{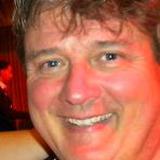} \\[-0.1cm]
    & & \parbox[t]{0.12\textwidth}{\centering\small Blurred} & & & & \\[0.1cm]

    \includegraphics[width=0.12\textwidth]{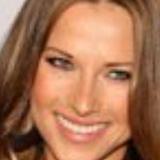} &
    \includegraphics[width=0.12\textwidth]{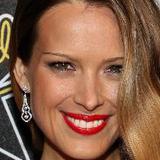} &
    \includegraphics[width=0.12\textwidth]{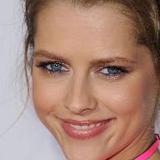} &
    \fcolorbox{red}{white}{\includegraphics[width=\dimexpr0.12\textwidth-2\fboxrule-2\fboxsep\relax]{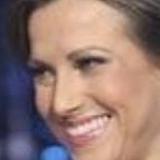}} &
    \includegraphics[width=0.12\textwidth]{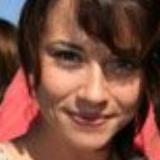} &
    \includegraphics[width=0.12\textwidth]{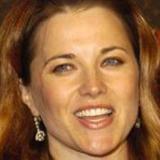} &
    \includegraphics[width=0.12\textwidth]{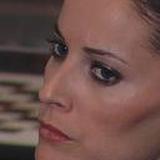} \\[-0.1cm]
    & & & \parbox[t]{0.12\textwidth}{\centering\small Pose} & & & \\[0.1cm]

    \includegraphics[width=0.12\textwidth]{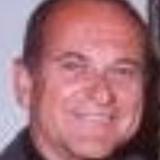} &
    \includegraphics[width=0.12\textwidth]{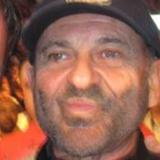} &
    \fcolorbox{red}{white}{\includegraphics[width=\dimexpr0.12\textwidth-2\fboxrule-2\fboxsep\relax]{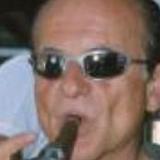}} &
    \includegraphics[width=0.12\textwidth]{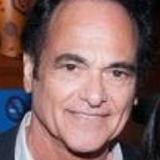} &
    \includegraphics[width=0.12\textwidth]{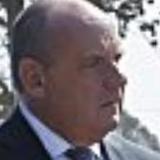} &
    \includegraphics[width=0.12\textwidth]{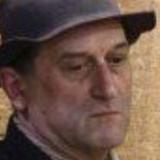} &
    \includegraphics[width=0.12\textwidth]{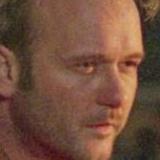} \\[-0.1cm]
    & & \parbox[t]{0.12\textwidth}{\centering\small Occlusion} & & & & \\[0.1cm]

    \includegraphics[width=0.12\textwidth]{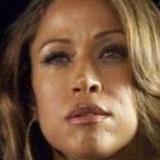} &
    \includegraphics[width=0.12\textwidth]{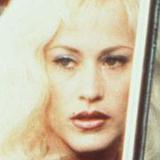} &
    \includegraphics[width=0.12\textwidth]{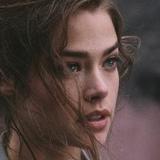} &
    \includegraphics[width=0.12\textwidth]{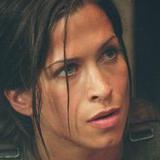} &
    \includegraphics[width=0.12\textwidth]{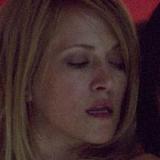} &
    \fcolorbox{red}{white}{\includegraphics[width=\dimexpr0.12\textwidth-2\fboxrule-2\fboxsep\relax]{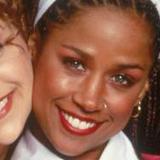}} &
    \includegraphics[width=0.12\textwidth]{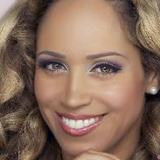} \\[-0.1cm]
    & & & & & \parbox[t]{0.12\textwidth}{\centering\small 2 Faces} & \\
    \end{tabular}

    \vspace{0.2cm}
    \caption{Facial recognition lineup examples showing one successful match (top row) and four failures. Each row displays a source image (left) followed by the constructed lineup. Green borders indicate successful recognition where the probe with the same identity as the source is correctly ranked first; red borders on the probe image indicate that the probe image was not correctly matched.}
    \label{fig:lineup_examples}
\end{figure*}
%
%

\subsection{Creating Lineups}
\label{ref:creating-lineups}

The increased scale of the dataset required some key changes to our evaluation framework, as our original implementation was not designed to efficiently handle such a large dataset.  On a single-GPU workstation, brute-force processing of the expanded dataset through the evaluation framework would take at least 21 hours per run. Under this configuration, running the entire accuracy pipeline -- including embedding extraction, similarity computation, lineup generation, and accuracy scoring” -- was computationally prohibitive.

The most substantial technical enhancement was the replacement of the brute-force similarity-based lineup generation process with Facebook AI Similarity Search (FAISS) for efficient embedding similarity computation~\cite{douze2024faiss}. Our original pairwise similarity approach scaled quadratically with dataset size, becoming computationally prohibitive as data volumes increased. FAISS addresses this through optimized indexing for high-dimensional vectors, enabling sub-linear similarity search through its IndexFlatIP structure. We also implemented L2-normalization on all embeddings and processed queries in batches of 256, allowing FAISS to compute similarities efficiently through vectorized operations. This optimization reduced overall processing time from 21 hours to approximately one hour per run, while maintaining comparable accuracy.

In the original evaluation framework~\cite{norman2023forensicfacialrecognition}, ArcFace achieved $82.3\%$ accuracy and FaceNet achieved $72.3\%$. When evaluated using our expanded framework with the full CASIA-WebFace dataset, ArcFace accuracy increased to $89.5\%$ while FaceNet accuracy remained comparable at $73.0\%$. Despite the substantially larger and more diverse dataset in the expanded framework, the overall accuracy levels remained fairly similar to the original evaluation. This consistency suggests that both models maintain robust performance across varying dataset sizes and compositions.

We note that Yi et al.~\cite{yi2014learning} report $97.73\%$ verification accuracy on LFW when training on CASIA-WebFace. This figure, however, represents a 1:1 verification on pre-cropped, aligned face pairs, a fundamentally different and easier task than our forensic lineup identification, which requires a 1:N ranking among adversarially-selected distractors.

\begin{figure}[t]
    \centering
    \includegraphics[width=0.9\columnwidth]{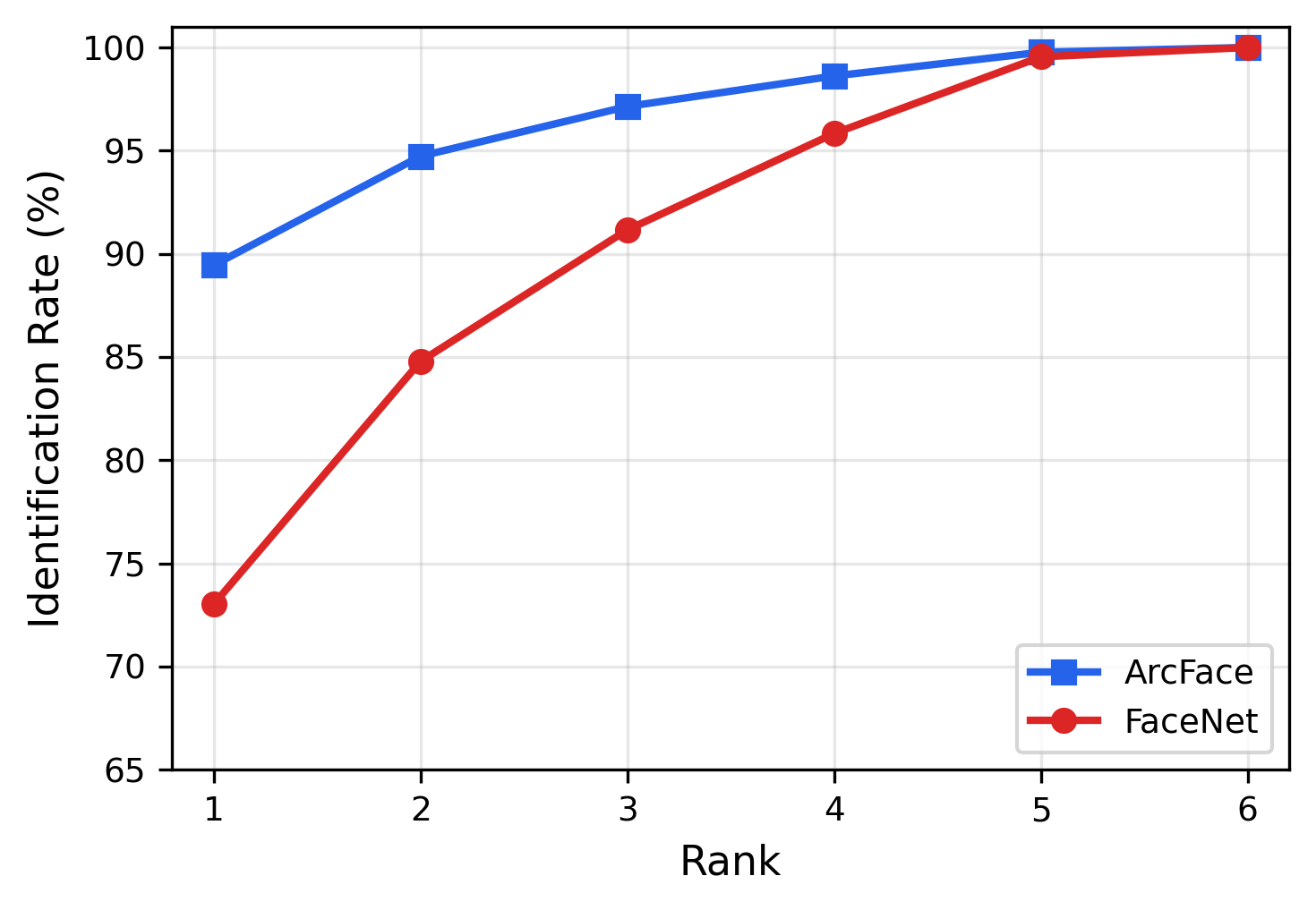}
    \caption{Cumulative Match Characteristic (CMC) for ArcFace and FaceNet on the expanded CASIA-WebFace framework. Both models reach $100\%$ accuracy within the 6-image lineup, but FaceNet shows substantially more rank-1 failures.}
    \label{fig:cmc_curves}
\end{figure}

Our new evaluation framework, combined with a larger dataset, allows us to begin to efficiently explore state-of-the-art methods for head-pose correction. In addition to rank-1 accuracy, we report Cumulative Match Characteristic (CMC) curves in Figure~\ref{fig:cmc_curves}. ArcFace reaches $94.7\%$ at rank-2 and $98.6\%$ at rank-4, while FaceNet rises from $73.0\%$ to $95.8\%$ at rank-4, indicating many FaceNet failures are near-misses.

Score margin analysis confirms this: correct identifications have a mean similarity margin of $+0.20$ over the top distractor, while failures average only $-0.05$, suggesting that modest restoration-driven improvements in probe similarity can convert a meaningful fraction of failures.

\begin{table}[t]
\centering
\caption{Lineup identification accuracy (\%) across restoration methods. ArcFace selective results are filtered through the failure prediction classifier (Section~\ref{ref:predicting-failure}); FaceNet selective represents the oracle upper bound (restoration applied to all true failures only, without a classifier gate). Universal denotes restoration applied to all lineup images indiscriminately.}
\label{tab:comparative_results}
\begin{tabular}{lrr}
\toprule
\textbf{Method} & \textbf{ArcFace} & \textbf{FaceNet} \\
\midrule
Baseline (expanded framework) & 89.5 & 73.0 \\
NextFace & 43.4 & 25.0 \\
CFR-GAN & 80.4 & 58.7 \\
NextFace + CodeFormer & 56.6 & 46.4 \\
CFR-GAN + CF (universal) & 77.8 & 68.7 \\
CFR-GAN + CF (selective$^\dagger$) & 92.8 & 84.4$^*$ \\
\bottomrule
\multicolumn{3}{l}{\footnotesize $^\dagger$ Classifier-gated ($91.6\%$ precision). $^*$ Oracle upper bound.}
\end{tabular}
\end{table}

\section{Head-Pose Correction}
\label{sec:head-pose-correction}

High-quality image reconstruction from a severely degraded or damaged source has emerged as a growing and crucial area of study within the field of computer vision. Contributions such as Krishnan et al.'s exploration of blind deconvolution to recover sharp images from blurred ones~\cite{krishnan2011blind} and Zamir et al.'s Restormer model~\cite{zamir2022restormer} are examples of technological and algorithmic improvements which have driven a major shift in image restoration.

As shown in our previous work, image enhancement models (even those tuned specifically for human body/face reproduction), can introduce both perceptually visible and invisible artifacts, which are then harmful to future recognition tasks~\cite{Norman_2024_CVPR}.

One of the main attributes that can make an image challenging for face recognition models is the subject's pose within the image. Front-facing images images make up the majority of most models' training data and, as such, are more likely to be successful in facial identification and recognition tasks.  However, most real-world data and datasets contain a high variance of head and body poses, which often prove challenging for facial recognition models.

In addition to pose, partial occlusion of faces is another challenge for facial recognition and identification models. Occlusion can occur as a result of subject position, scarves, hats, or sunglasses.  Any of these occlusions will result in less information available for feature extraction.

We evaluate two representative approaches spanning the primary paradigms for head-pose correction: NextFace, a 3D morphable model-based method that reconstructs facial geometry via differentiable ray tracing, and CFR-GAN, a 2D generative approach that performs joint frontalization and de-occlusion. While other approaches exist, including DR-GAN~\cite{tran2017drgan} and TP-GAN~\cite{huang2017tpgan}, our selection captures the fundamental distinction between explicit 3D reconstruction and learned 2D transformation. Our evaluation framework is model-agnostic and can readily accommodate additional correction methods.

\begin{figure}[t]
    \centering
    \small
    \setlength{\tabcolsep}{4pt}
    \renewcommand{\arraystretch}{1.05}

    \begin{tabular}{cc}
    \textbf{Input} & \textbf{NextFace Output} \\
    \midrule
    \includegraphics[width=0.45\columnwidth]{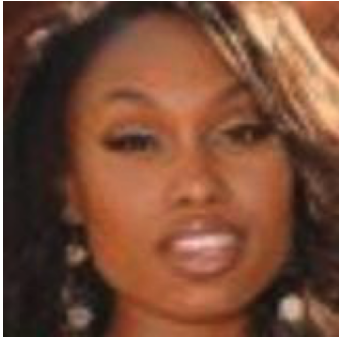} &
    \includegraphics[width=0.45\columnwidth]{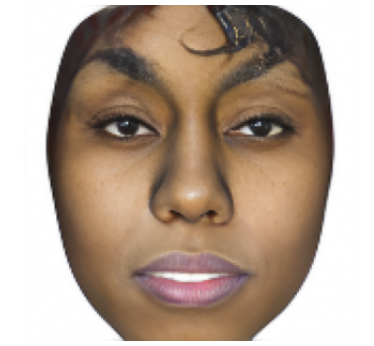} \\[0.2cm]
    \includegraphics[width=0.45\columnwidth]{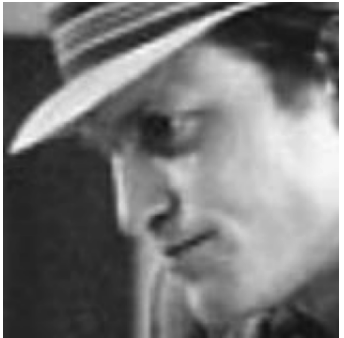} &
    \includegraphics[width=0.45\columnwidth]{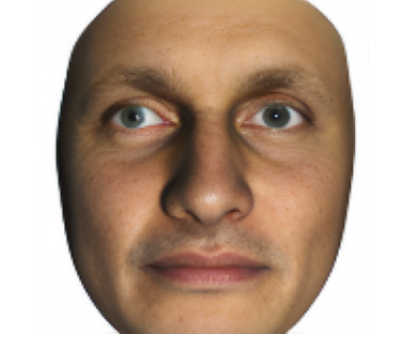} \\
    \end{tabular}

    \vspace{0.2cm}
    \caption{NextFace frontalization.}
    \label{fig:nextface_pipeline}
\end{figure}

\begin{figure}[t]
    \centering
    \small
    \setlength{\tabcolsep}{4pt}
    \renewcommand{\arraystretch}{1.05}

    \begin{tabular}{cc}
    \textbf{Input} & \textbf{CFR-GAN Output} \\
    \midrule
    \includegraphics[width=0.45\columnwidth]{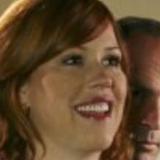} &
    \includegraphics[width=0.45\columnwidth]{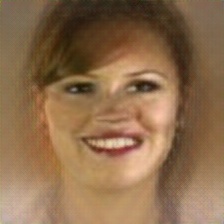} \\[0.2cm]
    \includegraphics[width=0.45\columnwidth]{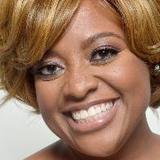} &
    \includegraphics[width=0.45\columnwidth]{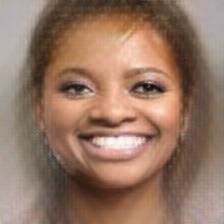} \\
    \end{tabular}

    \vspace{0.2cm}
    \caption{CFR-GAN frontalization.}
    \label{fig:cfrgan_pipeline}
\end{figure}

\subsection{3D Head Pose Correction (NextFace)}

In 2021, Dib et al.~\cite{Dib_2021_ICCV} introduced a self-supervised approach for monocular face reconstruction that integrates a CNN encoder with a differentiable ray tracer, capturing high-fidelity geometry and reflectance from single images. As shown in Figure~\ref{fig:nextface_pipeline}, this approach yields realistic morphable face models while maintaining near real-time inference performance.

\subsection{2D Head Pose Correction (CFR-GAN)}

Recently, generative computer-vision models, such as the Complete Face Recovery GAN~\cite{Ju_2022_WACV}, seek to ``rectify'' extreme-pose face images to a neutral frontal pose by effectively combining 2D face reconstruction, texture swapping, self-supervised learning, and generative adversarial training to simultaneously solve the challenging problem of joint face rotation and de-occlusion. As illustrated in Figure \ref{fig:cfrgan_pipeline}, this process results in high-quality, occlusion-free images that can be used for various downstream face-related tasks.

\subsection{Lineup Accuracy after Head-Pose Correction}

Surprisingly, the 2D and 3D restoration approaches led to a significant degradation in overall facial recognition accuracy. When lineup images were restored using CFR-GAN, accuracy was $58.7\%$ for FaceNet and $80.4\%$ for ArcFace. By comparison, baseline performance for FaceNet and ArcFace was $73.0\%$ and $89.5\%.$ When lineup images were restored using NextFace, accuracy declined even further to $25.0\%$ for FaceNet and $43.4\%$ for ArcFace. These head-pose correction techniques, while improving visual quality, appear to remove or distort distinctive facial features necessary for accurate facial recognition.

Visual inspection of the outputs reveals substantial artifacts, particularly for non-frontal inputs: CFR-GAN introduces blurring that homogenizes facial textures, while NextFace occasionally produces warped structures and omits neck/shoulder profiles (Figures~\ref{fig:nextface_pipeline},~\ref{fig:cfrgan_pipeline}). These artifacts motivated our investigation of a secondary restoration step.

\subsection{2D Restoration (CodeFormer)}

Since both NextFace and CFR-GAN resulted in the introduction of significant artifacts, we explored a secondary image restoration technique. Approaches such as the CodeFormer codebook-lookup transformer model~\cite{zhou2022towards} seek to increase the fidelity of low-quality images by passing a degraded image through a feature extraction module. These features are then processed by CodeFormer, which leverages a learned discrete codebook to represent high-quality facial attributes. The discrete codebook lookup helps reduce the ambiguity often found in mapping degraded inputs to restored outputs.

In our implementation, CodeFormer was configured with a controllable feature transformation weight of w $w=0.7$, balancing fidelity to the original image against enhancement quality. Face detection and alignment used RetinaFace. Restored outputs were resized to $160 \times 160$ pixels to match dataset dimensions.

\begin{figure}[t]
    \centering
    \small
    \setlength{\tabcolsep}{4pt}
    \renewcommand{\arraystretch}{1.0}

    \begin{tabular}{ccc}
    \parbox[t]{0.30\columnwidth}{\centering\textbf{Source}} &
    \parbox[t]{0.30\columnwidth}{\centering\textbf{CFR-GAN}} &
    \parbox[t]{0.30\columnwidth}{\centering\textbf{CFR-GAN + CF}} \\
    \midrule
    \includegraphics[width=0.30\columnwidth]{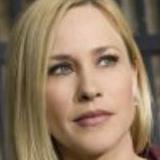} &
    \includegraphics[width=0.30\columnwidth]{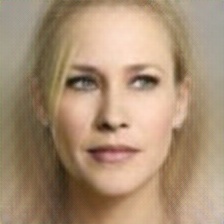} &
    \includegraphics[width=0.30\columnwidth]{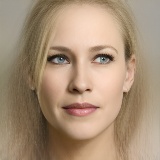} \\
    \\[-0.3cm]
    \midrule
    \parbox[t]{0.30\columnwidth}{\centering\textbf{Source}} &
    \parbox[t]{0.30\columnwidth}{\centering\textbf{NextFace}} &
    \parbox[t]{0.30\columnwidth}{\centering\textbf{NextFace + CF}} \\
    \midrule
    \includegraphics[width=0.30\columnwidth]{figures/cfrgan+CF/103_orig.jpg} &
    \includegraphics[width=0.30\columnwidth]{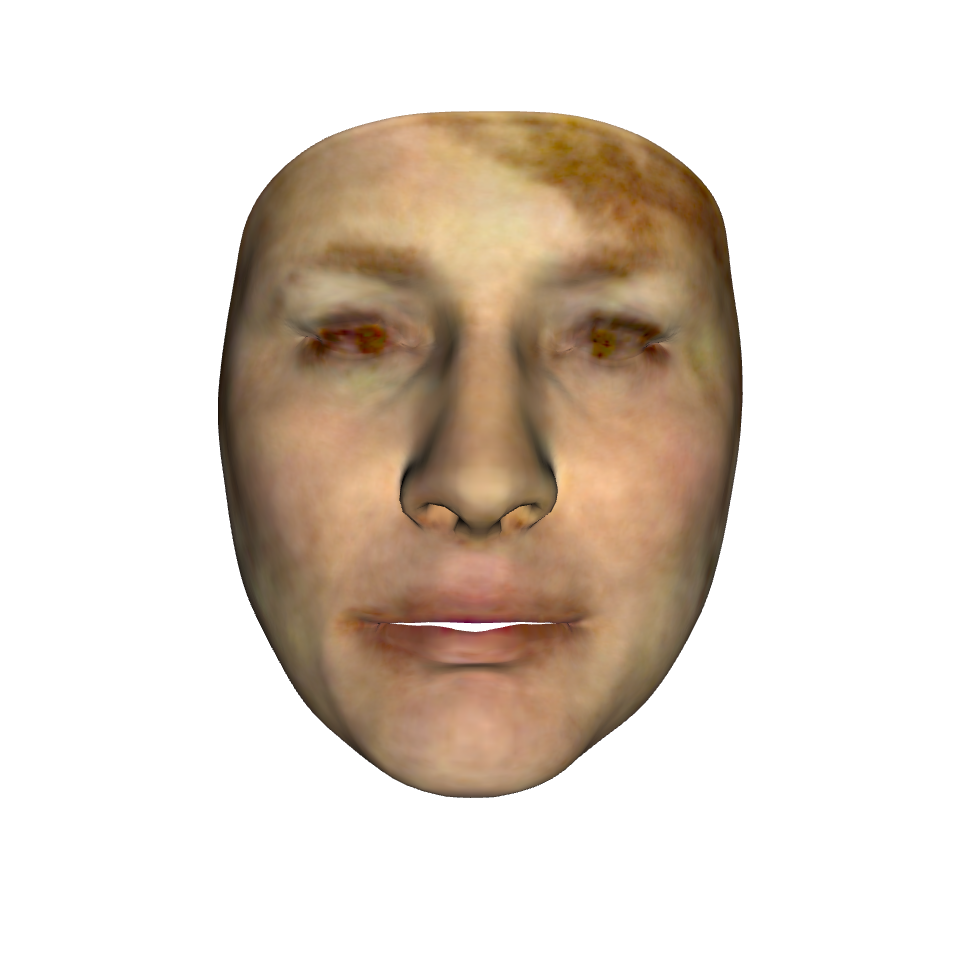} &
    \includegraphics[width=0.30\columnwidth]{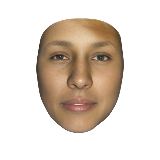} \\
    \end{tabular}

    \vspace{0.2cm}
    \caption{CFR-GAN frontalization followed by CodeFormer (CF) enhancement (top row) and NextFace frontalization followed by CF (bottom row).}
    \label{fig:restoration_pipeline}
\end{figure}

Shown in Figure~\ref{fig:restoration_pipeline} are examples of applying CodeFormer after head-pose estimation.

\subsection{Lineup Accuracy after Head-Pose Correction and Restoration}

In order to determine the effectiveness of restoration techniques on the evaluation framework, we created four new variants of our evaluation dataset. Each variant contained lineup images (excluding source images), which were processed through a different restoration approach: CFR-GAN alone, NextFace alone, CFR-GAN combined with CodeFormer, and NextFace combined with CodeFormer. We then applied the evaluation pipeline to these restored datasets.

Running the evaluation on NextFace combined with CodeFormer yielded $56.6\%$ for ArcFace and $46.4\%$ for FaceNet. While improved over NextFace alone ($43.4\%$ ArcFace, $25.0\%$ FaceNet), these results remain well below the expanded framework baselines (Table~\ref{tab:comparative_results}). As a result, we chose to discontinue use of this head-pose restoration technique.

The combination of CFR-GAN with CodeFormer, when applied universally to all lineup images, yielded $77.8\%$ for ArcFace and $68.7\%$ for FaceNet. For ArcFace, this represents a degradation of $-11.7$ percentage points from the expanded baseline, while FaceNet shows a comparable degradation of $-4.3$ percentage points. Both models exhibit reduced accuracy under universal restoration, confirming that indiscriminate application of restoration techniques harms recognition performance regardless of the underlying architecture. The two-stage approach (geometric correction through CFR-GAN followed by feature enhancement through CodeFormer's learned codebook), proved more effective than either technique alone. Given NextFace's poor standalone performance, we focused subsequent analysis on the CFR-GAN + CodeFormer pipeline.

We applied CFR-GAN + CodeFormer restoration to all 30,535 previously failed ArcFace lineups. Of these, 9,749 ($31.9\%$) achieved successful identification (probe at rank~0) after restoration. The remaining 20,786 lineups did not achieve rank~0; their rank change distribution is shown in Table~\ref{tab:rank_changes}. Among those 20,786 lineups, $43.0\%$ showed positive rank movement (improved probe position), $31.9\%$ were unchanged, and $25.1\%$ showed negative movement. The average rank change for non-converted lineups was $+0.53$ positions, indicating a net positive shift even among cases that did not achieve successful identification.

\begin{table}[t]
\centering
\caption{Probe rank changes after CFR-GAN + CodeFormer restoration, for the 20,786 ArcFace lineups where the probe did not achieve rank~0 after restoration. Of 30,535 total failed lineups processed, the remaining 9,749 achieved rank~0 (successful identification) and are not included. Positive values indicate improved probe ranking.}
\label{tab:rank_changes}
\begin{tabular}{r@{\hspace{0.5cm}}r@{\hspace{0.5cm}}r}
\textbf{Rank Change} & \textbf{Count} & \textbf{Percentage} \\
\midrule
$-5$ positions & 1 & 0.01\% \\
$-4$ positions & 443 & 2.13\% \\
$-3$ positions & 932 & 4.48\% \\
$-2$ positions & 1,530 & 7.36\% \\
$-1$ positions & 2,315 & 11.14\% \\
No change & 6,629 & 31.89\% \\
$+1$ positions & 2,816 & 13.55\% \\
$+2$ positions & 2,153 & 10.36\% \\
$+3$ positions & 2,072 & 9.97\% \\
$+4$ positions & 1,895 & 9.12\% \\
\end{tabular}
\end{table}

Positive rank changes occurred more frequently than negative ones ($43\%$ vs.\ $25.1\%$), demonstrating that CFR-GAN + CodeFormer restoration provides a net benefit even when it does not fully resolve the identification failure. However, these results assume oracle knowledge of which lineups have failed. In real-world deployment, ground truth is unavailable and restoration must be applied based on predicted failures, motivating the classifier described next.

\section{Predicting Failure}
\label{ref:predicting-failure}

Analysis of the results from the correction and restoration process establishes that CFR-GAN + CodeFormer can potentially improve accuracy. However, when this treatment is applied to successful lineup identifications, $60\%$ of those same lineup identifications fail. As such, operational deployment requires the ability to predict potential biometric lineup failures ahead of time.

\subsection{Dataset}
\label{ref:restoration-dataset}

We developed a machine-learning approach to predict lineup failure. Our model was trained on a subset of the original CASIA-WebFace dataset, processed through the forensic lineup task, consisting of 244,825 unique lineup images from failed lineup identifications. This smaller dataset was first curated using Dlib~\cite{king2009dlib} and MediaPipe's Holistic ~\cite{mediapipe} to remove images with unnatural body poses, total obfuscation of faces, and extreme brightness, darkness, or blur. The resulting dataset had a class imbalance with a $12.5\%$ lineup failure rate, as opposed to the overall dataset failure rate of $\sim 33\%$.  The initial training dataset split sizes were: 176,274 samples ($72\%$) for training, 19,586 samples ($8\%$) for validation, and 48,965 samples ($20\%$) for final evaluation.

To address the inherent class imbalance, we generate multiple rebalanced training datasets using controlled undersampling of the majority class. Precision-focused datasets employ conservative ratios ranging from 1.2:1.0 to 2.0:1.0 (success:failure), creating 10 distinct training sets with failure rates between 33\% and 45\%. Recall-focused datasets use more aggressive ratios from 0.7:1.0 to 1.1:1.0, yielding failure rates between 48\% and 59\%. Each rebalanced dataset maintains the original minority class samples while randomly sampling the majority class according to the target ratio. Deterministic random seeds ensure reproducible sampling across training runs.

\subsection{Feature Engineering}
\label{ref:feature-engineering}

Our lineup-failure prediction model combines deep-learning embeddings with traditional image quality metrics, capturing both semantic content and low-level visual characteristics. The deep-learning features consist of 768-D OpenCLIP embeddings extracted from ViT-B/32 pre-trained on LAION-400M.

Traditional descriptive image features contribute 42 dimensions across six categories: \textbf{Lighting} (6: grayscale mean/std, entropy, dark/bright ratios, Laplacian variance), \textbf{Quality} (7: local/global contrast, dynamic range, Michelson/RMS contrast), \textbf{Noise} (5: diagonal difference std, SNR, median filter residuals), \textbf{Sharpness} (6: Sobel gradients, Laplacian variance, FFT energy), \textbf{Texture} (2: local variance, edge density), and \textbf{Face geometry} (16: detection status, face area/centering, eye/mouth aspect ratios, bilateral symmetry, and head pose angles via dlib's~\cite{king2009dlib} 68-landmark predictor). All image operations use OpenCV, with SciPy for entropy computation.

All 810 features undergo standardization using StandardScaler~\cite{Pedregosa_Scikit-learn_Machine_Learning_2011} to ensure equal contribution across different scales, with NaN and infinite values replaced by bounded defaults (0.0) and extreme values clipped to $\pm 1e6$ to ensure numerical stability during model training.

\subsection{Model Architecture and Training}
\label{ref:model-architecture}

Our model employs a hybrid two-stage ensemble architecture designed to balance precision (the proportion of predicted failures that were actual failures) and recall (the proportion of actual failures that were correctly identified) through complementary model specialization. This design is motivated by the forensic deployment context, where the cost of a false positive far exceeds the cost of a false negative. A single classifier optimized for precision sacrifices recall, while one optimized for recall generates unacceptable false positive rates. The dual-cohort ensemble resolves this trade-off through geometric mean aggregation, achieving $91.6\%$ precision while maintaining $51.8\%$ recall. The approach trains separate cohorts of precision-focused and recall-focused classifiers on our previously mentioned strategically rebalanced datasets, then combines their predictions using geometric mean aggregation.

The ensemble comprises 20 base classifiers in two cohorts of 10: precision-focused models (higher regularization, conservative tree depths 3--8, minimal class weighting) and recall-focused models (lower regularization, deeper trees, aggressive class weighting up to 1:2.5). Each cohort includes Logistic Regression, Gradient Boosting, Extra Trees, Random Forest, and XGBoost variants trained on stratified splits ($72\%$/$8\%$/$20\%$). The ensemble aggregates predictions via geometric mean, and the classification threshold (0.42) was selected via grid search on the validation set to maximize a composite precision-recall score.

\subsection{Model Performance}
\label{ref:model-performance}

After validation, the model achieved $91.6\%$ precision (defined as the proportion of predicted failures that were actual failures) and $51.8\%$ recall (the proportion of actual failures that were correctly identified), yielding an F1-score of $0.66$. This conservative approach was deliberately designed to minimize false positives while maintaining sufficient sensitivity to detect genuine lineup failures.

\section{Restoration Impact Analysis}
\label{ref:restoration-impact}

After applying the model described in the previous section, the lineup associated with any source image predicted to fail was subjected to the complete CFR-GAN + CodeFormer restoration. We then compared the movement of the correct identity in lineup ranking before and after restoration.

Among 3,178 true positive predictions, 877 samples ($27.6\%$) showed rank improvements after restoration processing. Most significantly, 221 samples transitioned from lineup failures before restoration (rank $>$ 0) to perfect matches (rank $=$ 0) after restoration.

These results demonstrate meaningful restoration improvements. True positive cases, where the model successfully identified a lineup failure and the corresponding image was restored, showed an average rank improvement of 0.39 positions, with original rankings averaging 1.83 and improving to 1.44 after restoration. This represents a $21.3\%$ improvement in average ranking position, indicating substantial quality enhancement for correctly identified failure cases.

Next, we analyze the false positive predictions, where the model incorrectly identified an image as a failure when it belonged to a successful lineup. Among the 292 incorrectly predicted failures, 260 samples ($89\%$) maintained their original ranking after the restoration processing, and only 31 samples ($10.7\%$) led to an incorrect lineup match. In particular, the original rank of 1.00 increased marginally to 1.11 after the restoration. This minimal impact demonstrates that the restoration pipeline's conservative application causes limited harm when applied to already successful lineups.

The experimental results validate our failure prediction approach across multiple dimensions. The binary classification model successfully identifies lineup failures with high precision, while the restoration pipeline demonstrates measurable improvements for correctly identified cases. The $27.6\%$ improvement among true positives, combined with the minimal negative impact of false positives, establishes a favorable risk-reward trade-off for automated restoration intervention.

The success conversion rate of $7.0\%$ represents particularly valuable outcomes, as these cases transition from failed identifications to perfect matches. When combined with the broader improvement rate, the system demonstrates clear utility for enhancing restoration quality in production environments.

\subsection{Cross-Model Restoration Impact}
\label{ref:cross-model-impact}

To assess generalizability across recognition architectures, we repeated the full restoration pipeline using FaceNet. On the expanded CASIA-WebFace framework, FaceNet achieves a baseline rank-1 accuracy of $73.0\%$ across 256,562 lineups, lower than ArcFace's $89.5\%$, reflecting greater sensitivity to adversarial lineup construction.

When CFR-GAN + CodeFormer restoration is applied universally, FaceNet accuracy decreases to $68.7\%$ ($-4.3$ pp). This parallels ArcFace, where universal application decreases accuracy to $77.8\%$ ($-11.7$ pp), confirming that indiscriminate restoration degrades both architectures.

Among FaceNet failures, $42.1\%$ were converted to successful identifications after restoration---a higher conversion rate than ArcFace's $31.9\%$. However, universal application also degraded $21.5\%$ of previously successful FaceNet lineups, compared to $10.7\%$ for ArcFace false positives under classifier-gated application. Under oracle-selective application (restoring only true failures), FaceNet accuracy reaches $84.4\%$, a $+11.4$ pp improvement.

These results confirm that: (1)~the restoration framework produces consistent improvement patterns across architectures, with both models benefiting from selective but not universal application; (2)~the degradation rate under universal application further motivates the classifier-gated selective intervention demonstrated for ArcFace; and (3)~FaceNet's larger failure population offers greater absolute improvement potential under selective restoration.


\section{Discussion}
\label{sec:discussion}

Our results provide a nuanced answer to the question: {\em Can targeted, AI-driven head-pose correction and image restoration meaningfully improve facial recognition accuracy in forensic-style conditions?} At a high level, we find that the answer is conditional. Naively applied, modern restoration and frontalization tools systematically harm performance, sometimes catastrophically. However, when these same tools are combined carefully and deployed selectively, they can yield measurable and practically meaningful improvements. This duality highlights the need for a shift in how enhancement technologies are evaluated and operationalized in biometric pipelines, particularly in high-stakes forensic settings.

Our findings have several implications for the forensic use of facial recognition systems: (1) Agencies and vendors should be wary of marketing or deploying plug-and-play enhancement modules that are applied blindly to all images. Even when they produce visually attractive results, these tools can systematically degrade biometric performance; (2) Restoration methods must be evaluated in the context of the specific recognition systems and tasks with which they will be paired. Improvements in human-perceived image quality or benchmark restoration metrics do not guarantee better forensic performance; (3) Any use of automatic restoration in a forensic pipeline should be fully documented, including the specific models used, their configuration, and when they are applied. Downstream practitioners, courts, and auditors need to know whether an image has been algorithmically altered and why; and (4) Our failure-prediction scores could be exposed to human examiners as an additional piece of meta-evidence—e.g., flagging lineups as high-risk for recognition failure and suggesting when additional human scrutiny or supplementary evidence is warranted.

Several limitations warrant discussion: (1) Our evaluation uses a curated subset of CASIA-WebFace, which, while large (491,414 images, 10,575 identities) and diverse, represents primarily web-crawled celebrity imagery. Forensic scenarios involving surveillance footage (motion blur, extreme compression, low-light noise) or bodycam imagery (lens distortion, rapid motion) may exhibit different degradation patterns. Cross-dataset validation on IJB-C~\cite{maze2018ijbc} or MegaFace~\cite{kemelmacher2016megaface} would strengthen external validity; (2) While we evaluate two recognition architectures (ArcFace and FaceNet) and observe differing restoration responses across them (Section~\ref{ref:cross-model-impact}), more recent architectures including AdaFace~\cite{kim2022adaface} and proprietary commercial systems may respond differently to restoration artifacts. Restoration effects may also be demographic-dependent, a critical concern for forensic deployment that warrants dedicated study; (3) Our failure prediction classifier was trained and evaluated on ArcFace lineups. The FaceNet results (Section~\ref{ref:cross-model-impact}) demonstrate that restoration benefits transfer across models, but the classifier itself would require retraining for each target architecture, as failure patterns are architecture-dependent.

These limitations suggest caution in directly extrapolating our quantitative results to all real-world deployments. Instead, we view our framework and findings as a template for how such systems should be evaluated and stress-tested.




\clearpage
\bibliographystyle{ieee_fullname}
\bibliography{main}

\end{document}